\documentclass [sigconf]{acmart}

\AtBeginDocument{%
  }

\copyrightyear{2026}
\acmYear{2026}
\setcctype{by}
\acmConference[WSDM '26]{Proceedings of the Nineteenth ACM International Conference on Web Search and Data Mining}{February 22--26, 2026}{Boise, ID, USA}
\acmBooktitle{Proceedings of the Nineteenth ACM International Conference on Web Search and Data Mining (WSDM '26), February 22--26, 2026, Boise, ID, USA}
\acmDOI{10.1145/3773966.3777969}
\acmISBN{979-8-4007-2292-9/2026/02}

\usepackage{amsthm}
\usepackage{algorithm2e}
\usepackage{multirow}
\usepackage{mathtools}

\colorlet{tablerowcolor}{gray!10} 

\begin{document}

\title{ReToP: Learning to Rewrite Electronic Health Records for Clinical Prediction}


\author{Jesus Lovon-Melgarejo}
\email{jesus.lovon@irit.fr}
\orcid{0000-0001-6243-0864}
\affiliation{%
 \institution{University of Toulouse, IRIT}
 \city{Toulouse}
 \country{France}
}

\author{Jose G. Moreno}
\orcid{0000-0002-8852-5797}
\email{Jose.Moreno@irit.fr}
\affiliation{%
 \institution{University of Toulouse, IRIT}
 \city{Toulouse}
 \country{France}
}

\author{Christine Damase-Michel}
\orcid{0000-0001-5018-0108}
\email{christine.damase-michel@utoulouse.fr}
\affiliation{%
 \institution{Toulouse University Hospital}
 \city{Toulouse}
 \country{France}
}
\affiliation{%
 \institution{University of Toulouse, Inserm UMR 1295}
 \department{CERPOP‑SPHERE Team}
 \city{Toulouse}
 \country{France} 
}

\author{Lynda Tamine}
\orcid{0000-0002-3615-8032}
\email{Lynda.Tamine@irit.fr}
\affiliation{%
 \institution{University of Toulouse, IRIT}
 \city{Toulouse}
 \country{France} 
}

\renewcommand{\shortauthors}{Jesus Lovon-Melgarejo, Jose G. Moreno, Christine Damase-Michel, and Lynda Tamine}
\begin{abstract} 
Electronic Health Records (EHRs) provide crucial information for clinical decision-making. However, their high-dimensionality, heterogeneity, and sparsity make clinical prediction challenging. Large Language Models (LLMs) allowed progress towards addressing this challenge by leveraging parametric medical knowledge to enhance EHR data for clinical prediction tasks.
Despite the significant achievements made so far, most of the existing approaches are fundamentally task-agnostic in the sense that they deploy LLMs as EHR encoders or EHR completion modules without fully integrating signals from  the prediction tasks. This naturally hinders task performance accuracy. In this work, we propose  \textbf{Re}write-\textbf{To}-\textbf{P}redict (\textbf{ReToP}), an  LLM-based  framework that addresses this limitation through an end-to-end training of an EHR rewriter and a clinical predictor. To cope with the lack of EHR rewrite training data, we generate synthetic pseudo-labels using clinical-driven feature selection strategies to create diverse patient rewrites for fine-tuning the EHR rewriter. \textbf{ReToP} aligns the rewriter with prediction objectives using a novel \textit{Classifier Supervised Contribution (CSC)} score that enables the EHR rewriter to generate clinically relevant rewrites that directly enhance prediction. 
Our \textbf{ReToP} framework surpasses strong baseline models across three clinical tasks on MIMIC-IV. 
Moreover, the analysis of \textbf{ReToP} shows its generalizability to unseen datasets and tasks with minimal fine-tuning while preserving faithful rewrites and emphasizing task-relevant predictive features.
\end{abstract}

\ccsdesc[500]{Computing methodologies~Natural language processing}

\keywords{LLMs, Electronic Health Record (EHR), Clinical Prediction}


\maketitle

\section{Introduction}\label{intro} 

Clinical predictive models heavily rely on electronic health records (EHRs), which encode longitudinal patients' medical features (e.g., disease, procedures)  to estimate the risks of having or developing a health-related outcome.  Clinical prediction faces several challenges, among which are heterogeneity, high dimensionality, and sparsity of EHRs \cite{Landi_2020,Zhu24}. Significant research effort has been dedicated to tackle these challenges through the development of machine learning models for a wide-range of clinical prediction tasks including diagnosis prediction \cite{SUSHIL:2018,dligach-miller-2018-learning,hur2023genhpf,kim2024general}, mortality prediction \cite{Choi2020, SUSHIL:2018,hur2023genhpf, naik-etal-2022-literature}, readmission prediction \cite{jiang2024graphcare,hur2023genhpf} and length of stay prediction \cite{Stoneetal2022,hur2023genhpf}. 
One major agreed-upon finding from these studies is that leveraging expert knowledge with patient data insights enhances EHR modeling, which leads to significant improvement in  prediction accuracy. Previous studies mostly integrated expert symbolic knowledge in the form of clinical knowledge graphs that represent medical features (e.g., ICD10 codes of diagnoses) and their  relationships to model valuable contextual information to complement EHR data \cite{Choi:2018, Choi2020,Tamine22,xu2024ram}. 

\begin{figure}[tb]
    \centering
    \includegraphics[width=\columnwidth]{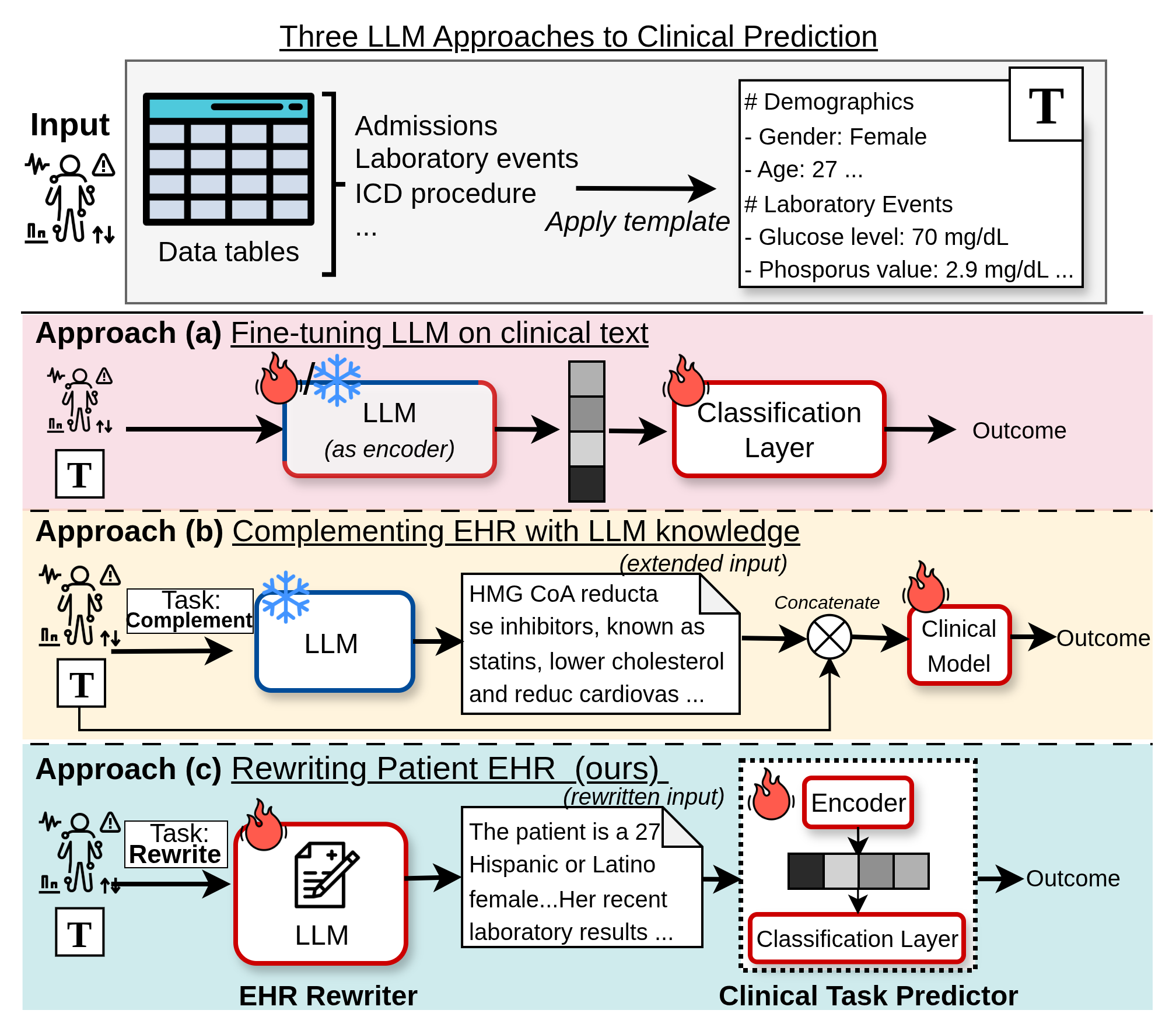}
    \caption{Comparison of three LLM-based approaches for clinical prediction from EHRs. Approach (a) relies on fine-tuned LLMs as EHR encoders. Approach (b) relies on LLMs to complement EHRs. Our approach, ReToP (c), uses a trainable  LLM-based EHR rewriter aligned with clinical tasks.}
    \Description{A three-layer comparison of different approaches to using LLMs for clinical prediction. Each layer illustrates the approach with the input elements and the model modules to use.}
    \label{fig:inference}
\end{figure}

Subsequently, Large Language Models (LLMs) have emerged as a significant milestone in the transition from symbolic knowledge to parametric knowledge, bringing us one step closer to potential knowledge graphs with impressive capabilities of language understanding and generation \cite{petroni-etal-2019-language}.  
To date, there is a large body of work showing the potential of LLMs to perform medical tasks \cite{NEURIPS2024_62986e0a,li-etal-2024-llamacare,kweon-etal-2024-publicly}. In the literature related to LLM-based clinical prediction, models rely on either of the following approaches illustrated in Figure \ref{fig:inference}: (a) leveraging fine-tuned pre-trained LLM  specialized on medical textual corpora  (e.g., Llamacare  \cite{li-etal-2024-llamacare}, Llemr  \cite{NEURIPS2024_62986e0a}). Although showing strong capabilities in interpreting complex EHR features, these works have been evaluated particularly on medical knowledge understanding tasks (e.g., medical question-answering)  and usually require a significant fine-tuning cost for achieving generalization ability \cite{Liu24};  (b) complementing EHRs with clinical kowledge from LLMs either agnostically \cite{xu2024ram} or dependently of EHR data before training the predictive model \cite{jiang2024graphcare}. However,  these existing works have two main limitations: (1) they have low flexibility in determining to what extent a certain knowledge is useful vs. noisy when naively fusing between LLM parametric knowledge and KG knowledge \cite{Chen_Lin_Han_Sun_2024,liu-etal-2024-knowledge-graph}; (2) they do not link EHR completion with clinical prediction. Since knowledge graphs contain a large amount of relational information, complementing EHR with noisy data may hinder the prediction accuracy.

Inspired by how human experts diagnose, we raise a critical question: "\textit{Given the patient EHR, what are the clinical rationales to support prediction?}" Our work answers this question while tackling the above-cited limitations.  We design an LLM-based EHR rewriter, acting as the unique source of expert knowledge, able to generate faithful and salient descriptions of the original patient EHR. The LLM-based EHR rewriter is further fine-tuned to revise the EHR rewrites to better support the clinical prediction. However, the  implementation of this answer poses the two following challenges:
\begin{itemize}
   \item \textbf{C1}: Given an input patient EHR, the LLM-based rewriter might be able to faithfully recall among a huge number of clinical features, the most salient ones, for any prediction task.     
   A potential solution is using an off-the-shelf LLM-based EHR rewriter. However, previous work showed that LLMs still struggle to comprehend tabular data, including EHRs \cite{Sui2024,singha2023tabular,lovon2025}. Since the gold relevant clinical features of an EHR are a priori unknown, the alternative of designing a trainable LLM-based EHR rewriter poses the challenge of a lack of gold supervision.  
   
    \item \textbf{C2}: The relevant clinical features of EHRs  are likely to be task-dependent. Thus, prediction accuracy might serve as a proxy evaluation of the quality of the EHR rewrite. However, accuracy can only be measured on gold task-specific  data, which do not involve patient EHR rewrites but original patient EHR instead. 

\end{itemize}

In this work, we introduce \textbf{Re}write \textbf{To} \textbf{P}redict (\textbf{ReToP}), a new LLM-based framework for clinical prediction.  As shown in Figure \ref{fig:inference} (approach c), \textbf{ReToP} is composed of two modules. The first module represents an LLM-based EHR rewriter, and the second module is a clinical task predictor.  Our framework is generalizable to different predictive clinical tasks at the light cost of fine-tuning the predictor without fully re-training the LLM-based rewriter.

To cope  with the high dimensionality and sparsity of clinical features and be able to generate faithful EHR descriptions (challenge \textbf{C1}), we train the EHR rewriter module using a set of synthetic pseudo-labels of patient-rewrites in two stages. 
We first sample EHR features to build diverse paraphrases to reduce dimensionality while  covering salient features of the original EHR. 
Then, we select the top-K paraphrased EHRs as pseudo-labels based on a relevance score that quantifies their clinical utility to a set of representative clinical tasks. 
The selected pseudo-labels are then used for fine-tuning the LLM-based EHR rewriter.
To align patient EHR rewrite quality to clinical prediction accuracy (challenge \textbf{C2}), we propose an end-to-end supervised training of the EHR rewriter and clinical predictor.
The clinical predictor iteratively provides supervision to the LLM-based EHR rewriter, which can then optimize its parameters to generate patient rewrites that help prediction. The predictor supervision quantifies the impact of each patient rewrite on prediction accuracy by using a Classifier Supervised Contribution (CSC) score inspired by the LM Supervised Retriever (LSR) measure \cite{shi2024replug}. 

We conduct extensive experiments across multiple EHR datasets, including MIMIC-IV \cite{johnson2023mimic}, eICU \cite{pollard2018eicu}, and EFEMERIS \cite{hurault2011drugs}, as well as a set of representative clinical prediction tasks:  mortality prediction, readmission prediction, length of stay prediction, and congenital malformation prediction. The results show that the \textbf{ReToP} framework outperforms  recent state-of-the-art baselines across these clinical tasks, achieving improvements up to $23\%$ in AUC-ROC score with a positive effect of each of its components. Furthermore, we show that our framework is generalizable to out-of-domain data and tasks with light computational cost during fine-tuning of the clinical predictor, without additional fine-tuning of the LLM-based EHR rewriter. 
Finally, qualitative analysis reveals that our EHR rewriter preserves faithfulness to the original EHR, and that KL alignment can emphasize serendipitous clinical features, not expected by experts, although valuable for accurate prediction.


\section{Related Work}

\subsection{Clinical prediction}

Clinical decision-making has made significant progress since deploying deep learning models on EHRs \cite{ocy068}. Early models have particularly tackled the challenge of domain-specific adaptation and demonstrated their ability to improve the effectiveness of a wide range of predictive clinical tasks, including health failure prediction \cite{Choi:2018,choi2016retain,zhang2021grasp}, mortality prediction \cite{Choi2020,SUSHIL:2018,hur2023genhpf,naik-etal-2022-literature}, and diagnosis prediction 
\cite{SUSHIL:2018,dligach-miller-2018-learning,hur2023genhpf,naik-etal-2022-literature,kim2024general}. 

The follow-up rise of pre-trained language models (LMs) has led to a vast amount of research investigating their capabilities to tackle the challenge of EHR data scarcity and clinical language understanding tasks.  Among these models, BEHRT \cite{7801}, MedBERT \cite{Rasmy2020MedBERTPC}, TransformEHR \cite{Yang2023TransformEHRTE} and ClinicalBERT \cite{clinicalbert} have particularly led to progress in disease prediction. Recently, REMED \cite{kim2024general} proposed a retrieval-based  filtering approach to address the sequence length limitations of LMs when processing EHR events.

In the same line of transformer-based models, but with more impressive transfer capabilities, decoder-only LLMs have emerged as promising tools for predictive healthcare. The first category of work focused on the design of foundation models on EHRs to tailor their inherent abilities of multi-tasking and reasoning to clinical applications \cite{NEURIPS2024_62986e0a,li-etal-2024-llamacare,kweon-etal-2024-publicly}. For instance, Llemr \cite{NEURIPS2024_62986e0a} is an instruction-tuned multimodal model based on Llama \cite{liu2023llava} that jointly encodes vectorized EHR event representations with natural language questions, evaluated on EHR-based question-answering tasks, and adapted for clinical prediction tasks, including mortality, readmission, length-of-stay, and diagnosis prediction. 
The second category of work integrates LLMs in the pipeline of a clinical predictive task by achieving various roles such as supporting clinical reasoning \cite{nguyen-etal-2024-carer,shi-etal-2024-ehragent} or completing EHR descriptions to improve the prediction \cite{Liu24,jiang2024graphcare,nguyen-etal-2024-carer,xu2024ram}. Especially,   Jiang et al. \cite{jiang2024graphcare}  propose a framework that relies on LLMs prompted to build a graph-based patient representation, which is then used to train a deep neural healthcare predictive model based on a graph neural network (GNN). 
Unlikely, Nuguyen et al. \cite{nguyen-etal-2024-carer} adopt a retrieval-augmented approach to enhance patient representations using multiple domain-specific resources. The authors leverage the LLMs to generate a local patient representation, a summary of knowledge resulting from a retriever queried with concepts from the EHR. This representation is then complemented by a patient-visit representation, and both are trained to achieve a predictive task. \\
Our work is significantly different in that we attempt to improve clinical prediction by fine-tuning, instead of zero-shot prompting, LLMs to generate the patient EHR rewrites without any external knowledge. Furthermore, instead of using task-agnostic enhanced EHR representations as input, we show how to train an LLM-based EHR rewriter  using clinical prediction outcomes as supervision.

\subsection{LLMs as feature selectors}

Recent work  has increasingly explored the application of LLMs as feature selectors \cite{choi2022lmpriors,jeong2025llmselect,li2025exploring}, leveraging their multi-task ability to  shift from traditional statistical methods toward more semantically-aware feature selection approaches. 
Early work \cite{choi2022lmpriors} 
introduced a prompt-based approach that formulates feature selection as a binary classification problem. The models generate ``Yes'' or ``No'' responses based on log-probability differences between these tokens to indicate feature importance given task descriptions, target features, and feature descriptions. LLM-Select \cite{jeong2025llmselect} extensively explored zero-shot and few-shot approaches across multiple paradigms, using score-based, rank-based, and dialogue-based methods to elicit feature relevance. 
Moreover, to address challenges in specialized domains where LLMs may have limited knowledge of domain-specific features, such as healthcare, RAFS \cite{li2025exploring} proposed a retrieval-augmented generation (RAG) approach that retrieves relevant feature descriptions from external knowledge sources.

However, these approaches mainly rely on feature descriptions, which require specialized domain knowledge that becomes resource-intensive when handling large feature sets. Additionally, few-shot approaches remain highly susceptible to prompt  variations \cite{sclar2024quantifying,pmlr-v139-zhao21c}. In contrast, our work leverages the LLM's internal knowledge through controlled generation by fine-tuning an LLM-based EHR rewriter that implicitly operates on feature importance derived from statistical feature selection heuristics applied to raw EHR descriptions, rather than relying on EHR feature descriptions.

\section{Background and Notations}
An EHR is an individual patient record of a sequence of visits. In this section, we introduce the notions and notations used in our work and formulate the clinical prediction task.

\subsection{Basic notions}
\begin{itemize}
    
\item \textbf{Patient EHR.} A patient's EHR $P_i$ is composed of demographic information and longitudinal representation of the health conditions recorded  during a sequence of hospital visits $V^1 \dots V^{\vert N_i \vert}$. Each visit $V^e$ includes a set of feature-value tuples $\mathcal{T}_i^e=\{(f^t_k,val^t_k)_i^e\}_{t=1}^{T}$ where $t = \{1..T\}$ are discrete timestamps during the visit $V^e$, and feature $f_k \in \mathcal{F}$ with  $\mathcal{F}=\{f_{1}, \dots f_{n}\}$ a reference set of $n$ features from different modalities (e.g., demographic, disease, medication).

\item \textbf{Clinical prediction task}. We consider standard healthcare prediction tasks $s\in \mathbb{S}$, where the input space is a population of patient EHRs $\mathbb{P}$ with each EHR $P_i \in \mathbb{P}$ in the form of tuples sampled from $\bigcup_{t=1}^T{\mathcal{T}_i^t} $ with $1..T$ discrete timestamps, and the output space is $\mathcal{Y}$ with $y_i\in \mathcal{Y}$ is a binary label, i.e., $y_i=\{0,1\}$. 
Given a patient EHR $P_i$ and a clinical task $s$, a predictive function $f$ outputs the clinical outcome $y_i$. Without loss of generality, we consider the mortality prediction, readmission prediction, length of stay prediction, and congenital malformation prediction tasks (§5.2).

\item \textbf{Patient EHR verbalisation.}
As done in previous work \cite{Hegselmann2022TabLLMFC,lovon2025, singha2023tabular,Sui2024}, a verbaliser function $v$ is used to convert EHR feature-value tuples of a patient $P_i$ (i.e., $P_i= \{ \bigcup_{t=1}^T{\mathcal{T}_i^j(t)} \}$)  into a natural language description $\mathcal{P}_i$.
 In this work, we applied markdown templates, verbalizing each tuple $(f_k, val_k)$ into a string ``- $f_k$: $val_k$''. We concatenate different feature modalities with headers ``\# \{feature name\}''.
\end{itemize} 

To simplify the notation,  we also refer to $f$ as the predictive function that maps a text-based patient EHR $\mathcal{P}_i=v(P_i)$ to the clinical outcome $y_i$.

\subsection{Problem statement and solution overview}
\textbf{ReToP} employs two trainable parametrized models: a rewrite model $\mathcal{M_\theta}$ and a task-specific predictor model $f_\phi$.
Formally, \textbf{ReToP} reframes the prediction function $f$ that supports the clinical task $s$, as a joint \textbf{Re}write ($\mathcal{M_\theta}$)-\textbf{To}-\textbf{P}redict ($f_\phi$) model that computes likelihood of outcome $y_i$ conditioned to patient EHR $\mathcal{P}_i^s$. To compute this likelihood, we  adopt an ensemble strategy as done in previous work \cite{shi2024replug}:
\begin{equation}
p(y_i^s \vert \mathcal{P}_i^s, Rw, \theta, \phi)=\sum_{\tilde{\mathcal{P}_i^s} \in Rw} p(y_i^s \vert \mathcal{P}_i^s \oplus \tilde{\mathcal{P}_i^s}, \phi) \times \delta (\tilde{\mathcal{P}_i^s} , \mathcal{P}_i^s)
\end{equation}

where $Rw$ includes all the possible rewrites of $\mathcal{P}_i^s$, limited in practice to $n_i$ rewrites (§4.2.2), $\oplus$ denotes the concatenation of two sequences, $p(y_i^s \vert \mathcal{P}_i^s \oplus \tilde{\mathcal{P}_i^s}, \phi)$ is the $\alpha$-weighted linear combination computed by $f_\phi$ for outcome $y_i^s$ conditioned on patient EHR $\mathcal{P}_i^s$ and corresponding rewrite $\tilde{\mathcal{P}}_i^s$,

\begin{equation}
p(y_i^s \vert \mathcal{P}_i^s \oplus \tilde{\mathcal{P}_i^s}, \phi) = \alpha \times p(y_i^s \vert \tilde{\mathcal{P}_i^s}, \phi)  + (1-\alpha) \times p(y_i^s \vert \mathcal{P}_i^s,\phi).
\end{equation}

$\delta(\tilde{\mathcal{P}}_i^s,\mathcal{P}_i^s)$, computed by $\mathcal{M_\theta}$, as  the probability $\mathcal{M_\theta}(\tilde{\mathcal{P}}_{i}^s\vert\mathcal{P}_i^s)$ of generating $\tilde{\mathcal{P}}_i^s$  as a rewrite of  $\mathcal{P}_i^s$,  
$\theta$ and $\phi$ are  model parameters trainable using task-specific training data $\mathcal{D}=\bigcup_{s \in \mathcal{S}} \mathcal{D}^s$ with $\mathcal{D}^s=\{(\mathcal{P}_i^s,y_i^s)\}_{1\leq i \leq m_s}$ of $m_s$ patients $\mathcal{P}_i^s$ sampled from population $\mathbb{P}^s \subset \mathbb{P}$ with corresponding clinical outcomes $y_i^s$.

We leverage the capabilities of LLMs for text generation \cite{Li_2024,mo2023learning} to support the patient rewriter model $\mathcal{M}_\theta$ and use any backbone classifier model to support the prediction function $f_\phi$.         

To tackle challenge \textbf{C1}, we build  a synthetic dataset $\mathcal{D}_{Rw}$ across tasks to fine-tune $\mathcal{M}_\theta$. $\mathcal{D}_{Rw}$ is composed of sets of $m_i$ pseudo labels of rewrites to each seed patient  $\mathcal{P}_i$, such as $\mathcal{D}_{Rw}=\{(\mathcal{P}_i,\tilde{\mathcal{P}_{ij}})\}_{\mathcal{P}_i \in \mathbb{P}, 1\leq j \leq m_i } $. $\mathcal{D}_{Rw}$ is built-upon a top-K selection strategy over candidate patient rewrites ${Rw}$ generated using paraphrasing-based methods over original patient EHR $\mathcal{P}_i$.

To tackle challenge \textbf{C2}, \textbf{ReToP} is trained over a training task-specific dataset $\mathcal{D}_{Pr}^{s}$ 
for end-to-end training of the patient rewriter $\mathcal{M_\theta}$ and the prediction function $f_\phi$ of task $s$, such as: \\$\mathcal{D}_{Pr}^s = \{( \mathcal{P}_{i}^s,\tilde{\mathcal{P}}_{ij}^s, y_i^s) _{j=1\dots n_i} \vert (\tilde{\mathcal{P}}_{ij}=\mathcal{M_\theta}(\tilde{\mathcal{P}}_{ij}^s\vert\mathcal{P}_i^s)  \wedge (\mathcal{P}_i^s,y_i^s) \in \mathcal{D}^s  \}$.
Formally, the training of \textbf{ReToP} based on Eq. (1) can be reframed as an optimization problem where we seek among all the patient candidate rewrites  $\tilde{\mathcal{P}_{ij}^s}$ generated from the original patient  $\mathcal{P}_i^s$ by  model $\mathcal{M_\theta}$, the patient rewrite $\tilde{\mathcal{P}}_{ij}^{*}$ that maximizes the expectation of the  clinical prediction task $s$ 
as follows:

\begin{equation}
\tilde{\mathcal{P}}_{ij}{*}=\underset{\tilde{\mathcal{P}_{ij}^s}}{\arg\max}\,\mathbb{E}_{\mathcal{D}_{Rw}}\lbrack\mathcal{M_\theta}(\tilde{\mathcal{P}}_{ij}^s\vert\mathcal{P}_i^s)\rbrack=\underset{\tilde{\mathcal{P}_{ij}^s}}{\arg\max}\,\mathbb{E}_{\mathcal{D}_{Pr}^s} \lbrack f_\phi(y_i^s \vert \tilde{\mathcal{P}}_{ij}^s)\rbrack
\end{equation}

\section{Methodology }

In this section, we describe the details of the procedure for building the synthetic training dataset ($\mathcal{D}_{Rw}$) used for fine-tuning the EHR rewriter (§4.1) and our end-to-end training procedure of the patient rewriter and the clinical prediction model (§4.2).

\subsection{Fine-tuning the EHR rewriter}

\subsubsection{Synthetic training dataset generation}
To enhance the LLM's ability to rewrite patient EHRs for clinical tasks, we deploy a paraphrasing-based generation approach  since it has been shown to favor diversity and faithfulness \cite{jiang-etal-2021,Li_2024}. Unlike previous work, we apply paraphrasing operators over the input patient EHR $P_i$ built upon feature-value tuples $\mathcal{T}_{i}^e=\{(f_k^t,val_k^t)_i^e\}_t$. We aim to reduce feature dimensionality and improve clinical feature diversity while remaining \textit{faithful} to the original patient EHR. 
Specifically, we consider a set of clinical tasks $s\in \mathbb{S}$ and we generate for each task and each seed patient $P_i$ sampled from population $s \mathbb{P}^s \subset \mathbb{P}^s$, a set of $K$  feature-based paraphrases $\{P_{ij} \}\,j=1\dots K$ using operators dedicated to EHR data.  
In our work, we design $K=8$ feature-based paraphrasing operators, inspired by previous work in prompt learning \cite{zhu_emerge,li2025exploring}, grouped by their feature selection strategy: 
    
    \textbf{Heuristic-based $\pi^h$}: we select clinical features based either on the temporal or value criteria given the importance of such criteria  agnostic to task-specific outcome: 1) $\pi^h_t$: selects the x\% more recent feature-value tuples such as $P_{i1}= \bigcup_{t=s}^{T_i}{\mathcal{T}_i^e(t)} $
    with $s=(1-x)*T_i$; 2) $\pi^h_v$ selects clinical features with abnormal values based on reference ranges
    such as $P_{i2}=\{(f_k^t,val^t_k)_i^e \in \mathcal{T}_{i}^e \vert (val^t_k < Min_k) \vee (val_k^t > Max_k)\}$ with $[ \,Min_k \dots Max_k]\,$ is the range reference of feature $f_k$.
    
    \textbf{Data-driven $\pi^d$}: we apply traditional feature selection methods that identify task-relevant features: 1) $\pi^d_{mi}$: filtering by mutual information \cite{lewis1992feature}, 2) $\pi^d_{mrmr}$: based on minimum redundancy maximum relevance selection \cite{ding2005minimum}, and 3) $\pi^d_{rfe}$: recursive feature elimination \cite{guyon2002gene}. Each method computes a score per feature, and we select the top $\mathsf{x}\%$ with the highest scoring features. Formally, for $j \in \{3,4,5\}$ corresponding to methods $\{\pi^d_{mi}, \pi^d_{mrmr}, \pi^d_{rfe}\}$ respectively, we define:
$P_{ij} = \{(f_k^t,val^t_k)_i^e \in \mathcal{T}_{i}^e \mid \text{score}_{j}(f_k^t, \mathcal{D}^s) \geq \mathsf{x}_s \}$, where $\text{score}_{j}(f_k^t, \mathcal{D}^s)$ denotes the score assigned to feature $f_k$ by method $j$ given the dataset $\mathcal{D}^s$, and $\mathsf{x}_s$ is the $x^{th}$ percentile threshold of all feature scores for task $s$.

    \textbf{Random-based $\pi^r$}: to favor the diversity of candidate clinical predictive features and values while reducing the dimensionality of the EHR, we design two operators:  1) $\pi^r_f$ randomly selects a sample of clinical features and then corresponding feature-value tuples: $P_{i6}=\{(f_k^t,val^t_k)_i^e \in \mathcal{T}_{i}^e \vert f_k^t \subset \mathcal{F} \}$; 2) $\pi^r_v$ randomly selects a subset of feature-value tuples from $P_i$ such as $P_{i7}=\{(f_k^t,val^t_k)_i^e \in \mathcal{T}_{i}^e $ 
    Finally, we consider the trivial case of the identity operator  3) $\pi^r_I$ that simply copies and pastes the original EHR: $P_{i8}=P_i$.

For each patient EHR $P_i$ verbalized into $\mathcal{P}_{i}$, we transform each of the corresponding paraphrases  in $\{P_{ij}\}_{j=1\dots 8}$ into a natural language patient description $\mathcal{P}_{ij}$ using the verbalizer $v$. 
For each patient $\mathcal{P}_{i}$, we generate $K$ input pairs
$(\mathcal{P}_{i}, \mathcal{P}_{ij})$. Finally, we obtain for each task $s$ the  dataset $Rw^s$ of candidate patient rewrites with 
$Rw^s =  \{(\mathcal{P}_i, \mathcal{P}_{ij})_{j=1}^K\ \vert \mathcal{P}_i=v(P_i), \mathcal{P}_{ij}=v(P_{ij}) \}_{P_i\in s\mathbb{P}^{s}}$.

\begin{algorithm}[tb]
\small
\caption{Fine-Tuning the EHR rewriter}
\label{alg:patient_sampling}

\KwIn{Task-specific rewrites $\{Rw^s\}$, scorer datasets $\{\mathcal{D}_{sub}^s\}$, 
      base LLM rewriter $\mathcal{M}_\theta^0$, selection percentile $k\%$, 
      number of rewrites per patient $K$}
\KwOut{Fine-tuned EHR rewriter $\mathcal{M}_\theta$}

$\mathcal{D}_{Rw} \gets \emptyset$\;
\ForEach{task $s \in \mathbb{S}$}{
    $Scorer^s \gets \mathrm{Train}(f_\phi,\mathcal{D}_{sub}^s)$ \tcp{\footnotesize{Train scorer}}
    
    $S^s \gets \{ Scorer^s(\mathcal{P}_{ij}) \mid (\mathcal{P}_i,\mathcal{P}_{ij}) \in Rw^s \}$ \tcp{\footnotesize{Compute scores}}
    
    $\tau_s \gets \min\{x \in S^s : \mathrm{rank}(x) \le |S^s|k/100 \}$ \tcp{\footnotesize{Fix threshold}}
    
    \ForEach{$(\mathcal{P}_i,\mathcal{P}_{ij}) \in Rw^s$}{
        $\tilde{\mathcal{P}}_i^{*} \gets \emptyset$ \tcp{\footnotesize{Initialize candidate list}}
        
        \If{$Scorer^s(\mathcal{P}_{ij}) \ge \tau_s$}{
            $\tilde{\mathcal{P}}_{ij} \gets \mathcal{P}_{ij}$\;
            $\mathcal{D}_{Rw} \gets \mathcal{D}_{Rw} \cup \{ (\mathcal{P}_i, \tilde{\mathcal{P}}_{i}^{*}) \}$\; 
        }
    }
}

$\mathcal{M}_\theta \gets \mathrm{Fine\mbox{-}tune}(\mathcal{M}_\theta^0,\mathcal{D}_{Rw})$\;

\Return $\mathcal{M}_\theta$\;

\end{algorithm}

\subsubsection{Fine-tuning the EHR rewriter} 
The Algorithm \ref{alg:patient_sampling} presents the pseudocode for building the synthetic training dataset $\mathcal{D}_{Rw}$ and fine-tuning  the patient rewriter $\mathcal{M_\theta}$ given a set of clinical tasks $\mathbb{S}$. For each pair $(\mathcal{P}_i, \mathcal{P}_{ij})\in Rw^s$, we evaluate the clinical relevance  of the candidate rewrite $\mathcal{P}_{ij}$ for task $s$ using a task-specific scorer $Scorer^s$. 
In our work, we consider $Scorer^s$  with the same architecture as our predictor model $f_\phi$, trained on $\mathcal{D}_{sub}^s$, a subset of task-specific training data from $\mathcal{D}^s$  involving patients with corresponding rewrites in $Rw^s$ and their corresponding labels from the original EHR, such as $    \mathcal{D}_{sub}^s=\{ (\mathcal{P}_{ij}, y_i) \mid (\mathcal{P}_{i},\mathcal{P}_{ij}) \in Rw^s, (\mathcal{P}_i, y_i) \in \mathcal{D}^s\}$. 
Our intuition is that $Scorer^s$ will assign higher predictive scores to informative rewrites while penalizing less relevant ones. For each patient $\mathcal{P}_i$, we select high-scoring rewrites as synthetic pseudo-labels into $\mathcal{D}_{Rw}$, as follows:

\begin{equation}
    \mathcal{D}_{Rw}=\bigcup_{s \in \mathbb{S}} \{ (\mathcal{P}_i, \mathcal{\tilde{P}}_{ij})_{1\leq j \leq m_i}  \mid Scorer_s(\mathcal{\tilde{P}}_{ij})  \geq \tau_s \}_{\mathcal{P}_i \in \mathcal{D}_{sub}^s}
\end{equation}

with $m_i \leq K$ is the number of high-quality rewrites selected as pseudo labels for patient $\mathcal{P}_i$, $\tau_s$ is the score threshold obtained by $Scorer^s$ to keep  the top-k\% rewrites. As a result, we generate the unified  dataset $\mathcal{D}_{Rw}$ containing filtered rewrites across the set $\mathbb{S}$ of clinical prediction tasks.\\
Finally, we fine-tune the patient EHR rewriter $\mathcal{M_\theta}$ on $\mathcal{D}_{Rw}$ using causal language modeling with a task-agnostic instruction.

\subsection{End-to-End training of the EHR rewriter and the clinical predictor}

In this section, we detail the calculation of the clinical prediction likelihood $f_\phi$ followed by the co-training methodology of the patient rewriter $\mathcal{M_\theta}$ and the clinical predictor $f_\phi$.

\subsubsection{Computing the clinical prediction likelihood}

The  \textbf{ReToP}  framework is trainable using task-specific data $\mathcal{D}^s=\{(\mathcal{P}_i^s,y_i^s)\}_{1\leq i \leq m_s}$ where $\mathcal{P}_i^s$ is sampled from population $\mathbb{P}^{s}$ and $y_i^s\in \mathcal{Y}$ is the corresponding binary label.
To enhance the clinical predictor within the rewrite-to-predict framework, we augment $\mathcal{D}^s$ with training examples $\mathcal{D}_a^s$ composed of rewrites from both paraphrase operators and our fine-tuned EHR rewriter, with all rewrites $\mathcal{P}_{ij}^s$ inheriting labels $y_i^s$ from their original EHR $\mathcal{P}_i^s$.
{\small
\begin{equation}
\begin{split}
    \mathcal{D}_a^{s} = &\{ (\mathcal{P}_{ij}^s, y_i^s) \mid (\mathcal{P}_{i}^s,\mathcal{P}_{ij}^s) \in Rw^s, (\mathcal{P}_i^s, y_i^s) \in \mathcal{D}^s\} \\
    &\cup \{ (\mathcal{M}_\theta(\mathcal{P}_{ij}|\mathcal{P}_{i}), y_i^s) \mid (\mathcal{P}_i^s, y_i^s) \in \mathcal{D}^s\}
\end{split}
\end{equation}
}

We design the clinical predictive function as an encoder classification model, optimized using Binary Cross-Entropy (BCE) loss: 

\begin{equation}
\mathcal{L}_C= - \sum_{(\mathcal{P}_i^s,y_i^s) \in \mathcal{D}^s_a} [y_{i} \log p(y_{i} \vert \mathcal{P}_i, \phi) + (1 - y_{i}) \log (1 - p(y_i \vert \mathcal{P}_i, \phi))]
\end{equation}

\subsubsection{Alignment of the EHR rewriter with clinical prediction}
\begin{figure}[tb]
    \centering
    \includegraphics[width=\columnwidth]{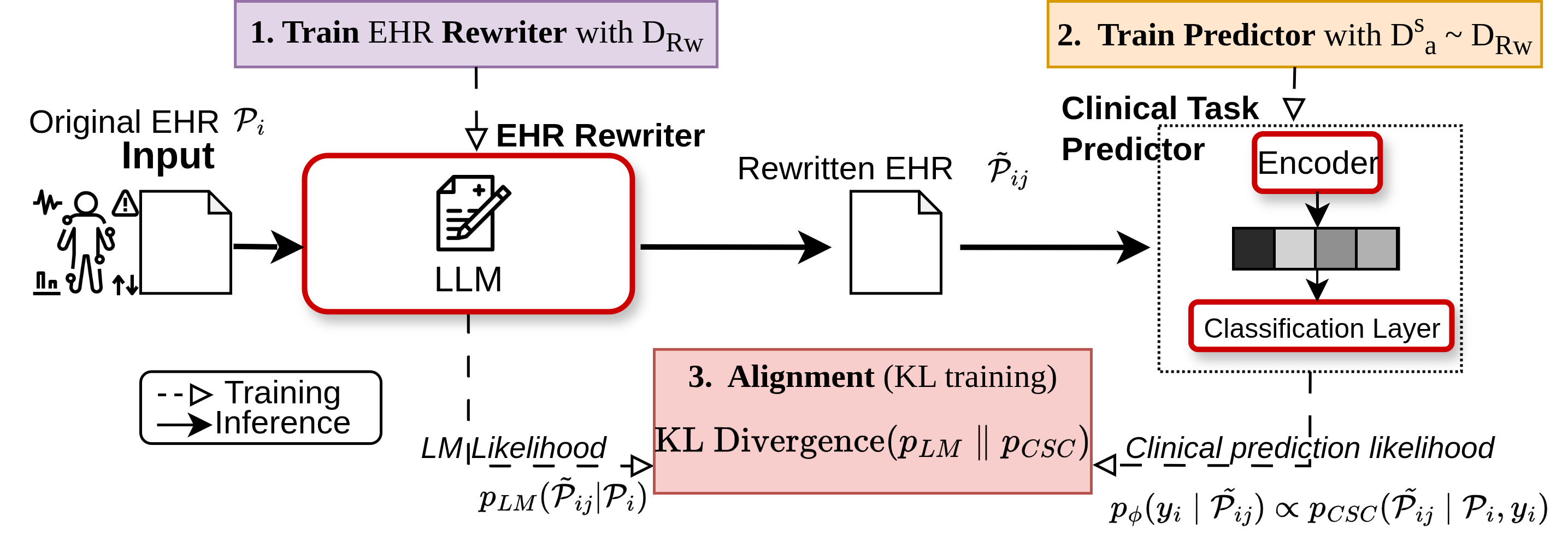}
    \caption{Overview of ReToP training and inference. We train the EHR rewriter on $\mathcal{D}_{Rw}$ (step 1). Then, jointly end-to-end train the rewriter and predictor on the target task (steps 2-3).}
    \Description{A diagram containing two modules from the ReToP framework. It shows the training and inference processes step by step, with corresponding equations that formalize each step.}
    \label{fig:training}
\end{figure}

As we formulate the overall clinical prediction task as an optimization problem  of the EHR rewriter (§3.2),  we propose a further fine-tuning of the EHR rewriter $\mathcal{M_\theta}$ to align its generation with the likelihood of the clinical predictor $f_\phi$, as presented in Figure \ref{fig:training}.  Specifically, we build a dual training dataset $\mathcal{D}_{Pr}^s=\{(\mathcal{P}_i^s,\tilde{\mathcal{P}}_{ij},y_i^s)_{j=1\dots n_i} \vert (\mathcal{P}_i^s,y_i^s) \in \mathcal{D}^s \wedge (\tilde{\mathcal{P}}_{ij}=\mathcal{M_\theta}(\tilde{\mathcal{P}}_{ij}\vert \mathcal{P}_i^s)  \} $. This fine-tuning aims to align the patient EHR rewriter with the clinical prediction performance, resulting in more relevant patient rewrite generation.
 Inspired by previous work \cite{sachan-etal-2023-questions,shi2024replug}, we introduce the  \textit{Classifier Supervised Contribution (CSC)} score, an adaptation of the LM-Supervised Retrieval (LSR) score \cite{shi2024replug}, for clinical prediction tasks. Given a training triplet $ (\mathcal{P}_{i}^s,\mathcal{\tilde{P}}_{ij},y_i^s) \in \mathcal{D}_{Pr}^s$, the CSC score quantifies the relative effectiveness of patient rewrite $ \tilde{\mathcal{P}}_{ij} $ to help accurately predicting  the target clinical outcome $y_i^s$ regarding all the candidate rewrites $\{\tilde{\mathcal{P}}_{ij}\}_{j=1}^{n_i}$ of patient $\mathcal{P}_{i}^s$:

\begin{equation}
    p_{CSC}(\mathcal{\tilde{P}}_{ij}\mid \mathcal{P}_i^s,y_i^s) = \frac{exp(p(y_i^s \mid \mathcal{\tilde{P}}_{ij},\phi) /\tau)}
    {\Sigma_{(\mathcal{P}_i^s,\tilde{\mathcal{P}}_{ij},y_i^s)\in \mathcal{D}_{Pr}^s} exp(p(y_i^s|\mathcal{\tilde{P}}_{ij},\phi) / \tau)} 
\end{equation}

 where $\tau$ is a temperature scaling parameter. The CSC score measures the contribution of a patient rewrite $\mathcal{\tilde{P}}_{ij}$ to the correct classification outcome $y_i^s$. To make  the patient rewriter model $\mathcal{M}_\theta$ guided by the clinical predictor $f_\phi$  we minimize the Kullback-Leibler (KL) divergence between the language model’s output distribution and the CSC-weighted probability:

\begin{equation}
    \mathcal{L}_{KL} = \mathbb{E}_{\mathcal{D}}\, KL \left( p_{LM}(\tilde{\mathcal{P}}_{ij}|\mathcal{P}_i^s,\theta) \parallel    p_{CSC}(\mathcal{\tilde{P}}_{ij} | \mathcal{P}_i^s,y_i^s) \right)  
\end{equation}

where $p_{LM}(\mathcal{\tilde{P}}_{ij}\mid \mathcal{P}_i^s,\theta)$ represents the rewriter's probability distribution over candidate rewrites, defined as:
\begin{equation}
    p_{LM}(\mathcal{\tilde{P}}_{ij}\mid \mathcal{P}_i^s,\theta) = \frac{exp(p(\mathcal{\tilde{P}}_{ij} \mid  \mathcal{P}_i^s,\theta) /\kappa)}
    {\Sigma_{(\mathcal{P}_i^s,\tilde{\mathcal{P}}_{ij},y_i^s)\in \mathcal{D}_{Pr}^s} exp(p(\mathcal{\tilde{P}}_{ij} \mid \mathcal{P}_i^s,\theta) / \kappa)} 
\end{equation}
with $\kappa$ as a temperature scaling factor.
The final training objective combines the KL loss with the standard language modeling loss:

\begin{equation}
    \mathcal{L}_{total} = \lambda \times  \mathcal{L}_{LLM} + (1 - \lambda) \times \mathcal{L}_{KL} 
\end{equation}

where $\mathcal{L}_{LLM}$ is the causal language modeling loss and $\lambda$ balances the two objectives. To complete the alignment process, we perform a final inoculation step of our clinical predictor $f_\phi$ on a small sample of rewrites generated by our KL-trained EHR rewriter.

\section{Experimental Design}

\subsection{Datasets}

We use three EHR datasets in our study: MIMIC-IV \cite{johnson2023mimic}, a publicly available dataset from Beth Israel Deaconess Medical Center. MIMIC-IV contains clinical data from hospital and ICU stays, and is widely adopted in healthcare research. To assess the generalization of our approach, we additionally consider two other datasets: eICU \cite{pollard2018eicu}, which gathers ICU data from many critical care units throughout the U.S, and EFEMERIS \cite{hurault2011drugs},  a private clinical dataset containing the medical history of pregnant women   as well as their neonatal outcomes regarding baby congenital malformation at birth.

\subsection{Clinical tasks and evaluation metrics}

\textbf{Mortality prediction (MOR)} predicts whether the patient will die in the next hospital visit, based on tuples from the current visit. Formally, $f: \{ \bigcup_{t=1}^{\mathcal{T}}{\mathcal{T}_i^e(t)} \} \mapsto y_i\lbrack \mathcal{T}_i^{e+1}\rbrack $ with $y_i\lbrack \mathcal{T}_i^{e+1}\rbrack \in \{0,1\}$ denotes the patient's mortality status.

\textbf{Readmission prediction (RA)} predicts if a patient will be readmitted into hospital within $15$ days.  Formally, $f: \{ \bigcup_{t=1}^{T}{\mathcal{T}_i^{e}(t)} \} \mapsto y_i\lbrack \mathcal{T}_i^{e+1}\rbrack$ with $y_i\lbrack \mathcal{T}_i^{e+1}\rbrack$ is equal to $1$ if $\Delta T(V^{e+1},V^e)\leq 15 $, $0$ otherwise, with $\Delta T$ is a time-interval function.

\textbf{Length-of-stay prediction (LOS)}  predicts whether the patient's hospital stay will be longer than $7$ days, using the first $48$ hours of the current visit. Formally, $f: \{ \bigcup_{t=1}^{\leq 48h}{\mathcal{T}_i^e(t)} \} \mapsto y_i\lbrack \mathcal{T}_i^e\rbrack$ with $y_i\lbrack \mathcal{T}_i^e\rbrack \in \{0,1\}$ denotes if the stay is longer than $7$ days.%

\textbf{Congenital malformation prediction (MALF)} predicts whether a newborn will present major malformation based on maternal EHR data. Formally, $f: \{ \mathcal{T}_i^1,\mathcal{T}_i^2,\mathcal{T}_i^3\} \mapsto y_i$, where $\mathcal{T}_i^j$  represents the maternal EHR data from month
$j$, and $y_i \in \{0,1\}$ indicates the presence of major malformation at birth.

We prepare task-specific cohorts as follows. For MOR and RA tasks, we omit patients  who are deceased in the current hospital admission.  For the LOS prediction task, we exclude patients with a length-of-stay shorter than 48 hours. Table \ref{tab:datasets} shows the final cohort statistics per task. 
We followed previous work pre-processing \cite{NEURIPS2024_62986e0a} for data preparation across all datasets. First, we exclude patients with more than two ICU stays per hospital admission, and with negative ICU or hospital length-of-stay, and patients under 18 years old. We then build our EHR by selecting the following tables: (1) for MIMIC-IV, we use hosp/patients, hosp/admissions, hosp/diagnosis, hosp/labevents, hosp/microbiologyevents, hosp/prescriptions, and hosp/transfers; (2) for eICU, diagnosis, microlab, lab, patient, medication, and treatment; and (3) EFEMERIS, demographics and prescriptions tables. We omit tables capturing dense bedside monitor signals or with substantial overlap with other tables\cite{johnson2023mimic,NEURIPS2024_62986e0a}. Finally, we randomly split each dataset into train, val, test sets (80/10/10). 

To evaluate prediction effectiveness,  we use the standard metrics for binary classification problems:  the Area Under the Receiver Operating Characteristic Curve (ROC) and the Area Under the Precision-Recall Curve (PRC) scores. We report mean scores and standard deviations computed via bootstrap sampling with replacement over $1000$ iterations.

\begin{table}[btp]
    \centering
    \small
        \caption{Cohort statistics for each task and dataset.}
    \begin{tabular}{llrrr}
    \toprule
    \textbf{Dataset}& \textbf{Task $(s)$} & \textbf{Size $(m_s)$}& \textbf{\# Neg.}& \textbf{\# Pos. ($\%$)}\\
    \midrule
    \multirow{3}{*}{MIMIC-IV}     &  MOR  &$29091$ & $28506$& $585 (2.01\%)$\\
    & RA & $29091$ &$12738$ & $16353 (52.21\%)$\\
    & LOS & $53771$ & $32637$ &$21134 (39.30\%)$\\
    \midrule
    \multirow{3}{*}{eICU} & MOR & $32750$& $31750$&$1000(3.05\%)$\\
    &RA&$32750$& $31840$&$910(2.78\%)$\\
    &LOS&$197174$&$185224$&$11950(6.06\%)$\\
    \midrule
    EFEMERIS & MALF&$134843$&$131623$&$3220 (2.39\%)$\\
    \bottomrule
    \end{tabular}
    \label{tab:datasets}
\end{table}

\subsection{Baselines}

We compare our approach with three main groups of baselines: (1) \textit{EHR-oriented models}, including classical methods such as RNN \cite{cho-etal-2014-properties}, RETAIN \cite{choi2016retain}, and GRASP \cite{zhang2021grasp}, which require particular hand-crafted features pre-processing following established protocols \cite{jiang2024graphcare,NEURIPS2024_62986e0a}. We also consider foundation models that can process raw EHR without particular pre-processing, including REMED \cite{kim2024general}, and Llemr \cite{NEURIPS2024_62986e0a}; (2) \textit{Serialized Classifiers}, which convert EHR data into text and fine-tune pre-trained language models with classification heads, including  ClinicalBERT \cite{clinicalbert}, Llama (Llama3-8B) \cite{dubey2024llama}, Qwen (Qwen 2.5-7b) \cite{qwen2}, and ModernBERT (base) \cite{modernbert}; (3) \textit{Rewrite then predict models}, which transform the original EHR input before classification. We evaluate all the proposed feature selection methods ($\pi^h$, $\pi^d$, $\pi^r$), including state-of-the-art ones $\pi^d_{mi}$ \cite{lewis1992feature}, $\pi^d_{mrmr}$ \cite{ding2005minimum}, and $\pi^d_{rfe}$  \cite{guyon2002gene} as well as self-gen \cite{Sui2024}, an LLM-based feature selector using Qwen2.5-7B as the backbone LLM.

\subsection{Implementation details}
\paragraph{Hardware and software configurations}
All training and evaluations are performed using CUDA 12.4, PyTorch 2.6.0 and the HuggingFace Transformers library. We train our models with 4 NVIDIA H100 80GB GPUs. To implement the baselines, we use PyHealth 1.1.6 framework \cite{pyhealth2023yang} when available. For efficient rewrite generation, we use the vLLM 0.8.3 library for accelerated inference. We perform grid search optimization for learning rates across all experiments, exploring values in the range $\{1e-3, 1e-4, 1e-5, 1e-6\}$.

\paragraph{Feature selector ($\pi$)} 
We implement the data-driven feature selection methods $\pi^d$ using the scikit-learn feature selection library. For the heuristic-based approaches $\pi^h$, we rely on the available clinical normal range values provided within each dataset. Across all approaches, following \cite{jeong2025llmselect}, we select the top-$x=30\%$ of features based on their computed relevance scores, and we add missing features among the top 10, to complement EHR information.

\paragraph{EHR rewriter} We evaluate our framework using two recent instruction tuned LLMs: Llama3-8B and Qwen2.5-7B. We construct our $\mathcal{D}_{Rw}$ dataset by fixing a top-$k=25\%$ for the quality threshold $\tau_s$. 
We fine-tune the rewriter using LoRA \cite{hu2022lora} with learning rate $2e-4$, $3$ training epochs, rank $r=8$, LoRA $\alpha=16$, dropout rate of $0.05$, bfloat16 precision, weight decay of $0.1$, batch size of $4$ and gradient accumulation steps of $4$. 
For KL fine-tuning, we construct $\mathcal{D}_{Pr}^s$ using $n_i=8$ rewrites per patient and train up to $4000$ steps using batch size $16$,  learning rate $2e-6$, $\kappa=0.01$, $\tau \in \{0.01, 0.1, 0.2, 0.3\}$ , and $\lambda \in \{0, 0.25, 0.5, 0.75\}$. Optimal parameters are selected based on task-specific evaluation every $1000$ steps.

\begin{table*}[tb]
    \centering
     \setlength{\tabcolsep}{3pt} 
    \small
    \caption{Results on the MIMIC-IV clinical tasks. \textbf{Bold} and \underline{Underline} show best and 2nd best scores. Metrics multiplied by 100.} 
    \begin{tabular}{lcccccc}
    \toprule
    \textbf{Model} &  \multicolumn{6}{c}{\textbf{MIMIC-IV}}\\
    \midrule
         & \multicolumn{2}{c}{\textbf{MOR}} &  \multicolumn{2}{c}{\textbf{RA}} & \multicolumn{2}{c}{\textbf{LOS}}    \\
         \cmidrule(lr){2-3} \cmidrule(lr){4-5} \cmidrule(lr){6-7}
         & \textbf{ROC}$(\uparrow)$ & \textbf{PRC}$(\uparrow)$ & \textbf{ROC}$(\uparrow)$  & \textbf{PRC}$(\uparrow)$& \textbf{ROC}$(\uparrow)$  & \textbf{PRC}$(\uparrow)$  \\
    \midrule
    &\multicolumn{6}{c}{\textit{EHR-oriented models}} \\
    
(a) RNN \footnotesize{(Choi et al., 2014 \cite{cho-etal-2014-properties})}&$63.33\pm 3.4$ & $3.38\pm 0.7$ &$65.76 \pm 1.2$ & $72.49 \pm 1.3$ & $73.02 \pm 0.7$ & $58.01 \pm 1.3$ \\
(b) RETAIN \footnotesize{(Choi et al., 2016 \cite{choi2016retain})} &	$59.13 \pm 3.9$ & $4.06 \pm 1.8$ & $64.57 \pm 1.2$ & $70.74 \pm 1.4$ & $72.91 \pm 0.7$ & $58.21 \pm 1.3$ \\
(c) GRASP \footnotesize{(Zhang et al., 2021 \cite{zhang2021grasp})} & $58.94 \pm 3.6$ & $2.68 \pm 0.5$ & $63.95 \pm 1.2$ & $68.92 \pm 1.4$ &$70.48 \pm 0.8$ & $53.85\pm 1.3$ \\
(d) Llemr \footnotesize{(Wu et al, 2024 \cite{NEURIPS2024_62986e0a})} & $52.07 \pm 4.1$ & $3.17 \pm 1.3$ &  $59.20 \pm 1.1$ & $64.89 \pm 1.3$ & $75.36 \pm 0.7$ &$62.21 \pm 1.3$  \\
(e) REMed \footnotesize{(Kim et al., 2024 \cite{kim2024general})} & $52.40 \pm 3.0$ & $2.16 \pm 0.5$ & $68.26 \pm 1.3$ & $75.31 \pm 1.9$& $80.15 \pm 1.9$ & $68.06 \pm 2.9$ \\
\midrule
&\multicolumn{6}{c}{\textit{Serialized Classifiers}} \\
(f) BioMedBERT \footnotesize{(Gu et al., 2020 \cite{pubmedbert})} 	&$64.24 \pm 3.6$	&$3.38 \pm 0.7$	&$70.36 \pm 1.0$&	$76.66 \pm 1.0$ & $77.59 \pm 0.7$	&$64.42 \pm 1.3$\\
(g) Qwen \footnotesize{(Yang et al., 2024 \cite{qwen2})}& $53.87 \pm 4.1$&$3.12 \pm 0.9$ & $61.06 \pm 1.1$&$66.15 \pm 1.3$&$73.78 \pm 0.7$&$60.75 \pm 1.3$\\
(h) Llama \footnotesize{(Dubey et al., 2024 \cite{dubey2024llama})}&$59.45 \pm 4.0$&$2.86 \pm 0.6$&$64.93 \pm 1.1$&$70.39 \pm 1.3$&$75.09 \pm 0.7$ & $62.42 \pm 1.2$ \\
(i) ModernBERT \footnotesize{(Warner et al, 2024 \cite{modernbert})}&	$58.01 \pm 3.7$&	$3.26 \pm 1.2$&	$70.98 \pm 1.0$&	$77.25\pm 1.0$&  \underline{$80.40 \pm 0.6$}&	$68.42 \pm 1.2$ \\
\midrule
&\multicolumn{6}{c}{\textit{Rewrite-then-Predict models}} \\
(j) self-gen \footnotesize{(Sui et al, 2024 \cite{Sui2024})} & $58.38 \pm 3.6$ &$2.95 \pm 0.9$ & $69.19 \pm 1.0$ & $76.24 \pm 1.0$ & $79.02 \pm 0.6$ &$67.04 \pm 1.2$\\
(k) $\pi^h_t$ & $61.00 \pm 4.2$ & $3.58 \pm 1.0$ & $65.25 \pm 1.0$ & $73.77 \pm 1.1$ & $70.04 \pm 0.7$ & $54.74 \pm 1.3$ \\
(l) $\pi^h_v$ & $56.05 \pm 3.7$ & $2.57 \pm 0.5$ & $64.47 \pm 1.0$ & $70.06 \pm 1.3$ & $66.63 \pm 0.8$ & $50.79 \pm 1.3$ \\
(m) $\pi^d_{mi}$ \footnotesize{(Lewis et al, 1992 \cite{lewis1992feature})} & $56.39 \pm 3.8$ & $2.93 \pm 0.7$ & $67.98 \pm 1.0$ & $74.57 \pm 1.1$ & $76.13 \pm 0.7$ & $60.57 \pm 1.2$ \\
(n) $\pi^d_{mrmr}$ \footnotesize{(Ding et al, 2005 \cite{ding2005minimum})} & $60.00\pm 3.6$ & $3.21 \pm 0.8$ & $63.08 \pm 1.1$ & $68.10 \pm 1.3$ & $79.48\pm 0.6$ & $68.53\pm 1.2$ \\
(o) $\pi^d_{rfe}$ \footnotesize{(Guyon et al, 2002 \cite{guyon2002gene})} & $62.54 \pm 3.5$ & $3.69 \pm 1.3$ & $62.03 \pm 1.0$ & $67.55 \pm 1.3$ & $76.84\pm 0.7$ & $63.22 \pm 1.2$ \\
(p) $\pi_f^r$ & $55.82 \pm 3.7$ & $2.63 \pm 0.7$ & $66.60 \pm 1.0$ & $72.34 \pm 1.2$ & $73.33 \pm 0.7$ & $59.29 \pm 1.3$ \\
(q) $\pi_v^r$ & $59.37 \pm 3.8$ & $3.73 \pm 1.5$ & $68.03 \pm 1.0$ & $74.37 \pm 1.1$ & $77.51 \pm 0.7$ & $63.41 \pm 1.2$ \\
\midrule
ReToP$_{\text{Qwen}}$ \footnotesize{\textit{(ours)}} & 
    \underline{$69.75 \pm 3.8$} & \underline{$4.61 \pm 1.0$} &\underline{$71.45 \pm 0.9$} &\underline{$77.92 \pm 1.0$}&$80.38 \pm 0.6$ & \underline{$68.91 \pm 1.2$}\\
ReToP$_{\text{Llama}}$ \footnotesize{\textit{(ours)}} & $\mathbf{71.92 \pm 3.6}$& $\mathbf{6.20 \pm 2.2}$&$\mathbf{72.05 \pm 0.9}$&$\mathbf{78.25 \pm 1.0}$&$\mathbf{80.62 \pm 0.6}$&$\mathbf{69.05 \pm 1.2}$ \\
\bottomrule
\end{tabular}
\label{tab:all_results}
\end{table*}

\paragraph{Clinical predictor} We employ identical hyperparameters for both the Scorer$^s$ classifier and clinical predictor. We sample $20\%$ of training patients to build $\mathcal{D}^s_{sub}$ and generate $\mathcal{D}_a^s$ with $3$ additional rewrites per patient. Both models use ModernBERT-base as the encoder backbone with a classification head, learning rate $2e-5$, context length $8192$ tokens, $10$ training epochs with early stopping patience $3$, batch size $32$, and gradient accumulation steps $4$. For the inoculation, we reduce the learning rate to $2e-6$  and $4000$ input samples to ensure optimal integration with the rewriter outputs.

\section{Experimental Results and Analysis}

\subsection{Main results}
Overall, Table \ref{tab:all_results} shows that the \textbf{ReToP} framework achieves a significant performance increase (t-test)  over all the baselines and across all the clinical tasks. The performance increase ranges are respectively up to $38.12\%$, $33.51\%$, and $28.84\%$ over respectively the \textit{EHR-oriented}, \textit{Serialized classifiers} and \textit{Rewrite-Then-Predict} baseline models.
Notably, we can see that among the \textit{Rewrite-then-predict models}, all the feature selector operators ($(k)\dots (q)$) do not show consistent performance improvements. In contrast, our framework with the \textbf{ReToP}$_{Llama}$ model achieves the highest performance across all the tasks, with improvements up to $+23\%$ compared to the best-performing baseline (ModernBERT). Specifically, \textbf{ReToP}$_{Llama}$ obtains improvements of  $13.92$ for MOR, $1.07$ RA, and $0.22$ for LOS tasks. \textbf{ReToP}$_{Qwen}$ shows similar improvements, demonstrating the robustness  of our framework. 

Finally, performance analysis across all the tasks raises an important observation:  imbalanced tasks such as MOR with only $2\%$ positive cases (§ Table \ref{tab:datasets}), leverage greater benefit from the \textbf{ReToP} framework ($+23\%$) compared to more balanced tasks like RA and LOS  tasks with larger training sets (with nearly twice the size of RA and MOR sets). 
This trend suggests that our rewrite-based approach provides greater value for rare clinical events, where high-quality synthetic representations can effectively address data scarcity, which is a common challenge in clinical prediction tasks.

\begin{table*}[bth]
     \centering
     \setlength{\tabcolsep}{1pt} 

    \footnotesize
    \caption{Ablation study results. \textbf{Bold} and \underline{Underline} indicate the best and 2nd best performance respectively. Values in parentheses () indicate percentage degradation relative to the complete ReToP framework, if any. All metrics are multiplied by 100.} 
    \begin{tabular}{@{}l*{12}{l}@{}}
    \toprule
    & \multicolumn{6}{c}{\textbf{Qwen2.5-7B}} & \multicolumn{6}{c}{\textbf{Llama3-8B}} \\
    \cmidrule(lr){2-7} \cmidrule(lr){8-13}
    
    & \multicolumn{2}{c}{\textbf{MOR}} & \multicolumn{2}{c}{\textbf{RA}} & \multicolumn{2}{c}{\textbf{LOS}} &
    \multicolumn{2}{c}{\textbf{MOR}} & \multicolumn{2}{c}{\textbf{RA}} & \multicolumn{2}{c}{\textbf{LOS}} \\
    \cmidrule(lr){2-3} \cmidrule(lr){4-5} \cmidrule(lr){6-7} 
    \cmidrule(lr){8-9} \cmidrule(lr){10-11} \cmidrule(lr){12-13}
    
    \textbf{Scenario} & \textbf{ROC} & \textbf{PRC} & \textbf{ROC} & \textbf{PRC} & \textbf{ROC} & \textbf{PRC} & \textbf{ROC} & \textbf{PRC} & \textbf{ROC} & \textbf{PRC} & \textbf{ROC} & \textbf{PRC} \\
    \midrule

w/o $\mathcal{D}_{Rw}$ &$64.92(6.9\%)$&$\mathbf{5.22}$&\underline{$69.49$}$(2.7\%)$&\underline{$76.11$}$(2.3\%)$&$78.78(2.0\%)$&$66.42(3.6\%)$& \underline{$70.03$}$(2.6\%)$ & $4.48(27.7\%)$ & \underline{$68.88$}$(4.4\%)$ & \underline{$75.38$}$(3.7\%)$ & $76.98(4.5\%)$ & $65.08(5.7\%)$\\
    w/o Rewriter&$58.83(15.7\%)$ &$3.16(31.5\%)$&$68.69(3.9\%)$&$74.67(4.2\%)$&\underline{$79.06$}$(1.6\%)$&\underline{$67.18$}$(2.5\%)$&$65.42(9.0\%)$& \underline{$4.93$}$(20.5\%)$&$68.86(4.4\%)$&$73.48(6.1\%)$&\underline{$79.37$}$(1.6\%)$&$67.27(2.6\%)$\\
    w/o KL & \underline{$64.93$}$(6.9\%)$    & $3.16(31.5\%)$ &$67.22(5.9\%)$   &$74.25(4.7\%)$  & $78.29(2.6\%)$ &   $66.44(3.6\%)$ & $59.00(18.0\%)$   & $3.61(41.8\%)$ &$67.40(6.5\%)$&$73.33(6.3\%)$&$79.09(1.9\%)$&\underline{$67.38$}$(2.4\%)$\\
    \midrule
    ReToP &  $\mathbf{69.75}$ & \underline{$4.61$} &$\mathbf{71.45}$ &$\mathbf{77.92}$&$\mathbf{80.38}$ & $\mathbf{68.91}$ &  $\mathbf{71.92}$& $\mathbf{6.20}$&$\mathbf{72.05}$&$\mathbf{78.25}$&$\mathbf{80.62}$&$\mathbf{69.05}$\\
    \bottomrule
    \end{tabular}
    
    \label{tab:ablation}
\end{table*}

\subsection{Ablation study}

We conduct comprehensive ablation studies to analyze the impact of different components of our \textbf{ReToP} framework at the training stage, with the following scenarios:
\begin{enumerate}
    \item \textit{w/o $\mathcal{D}_{Rw}$}: we replace our synthetic dataset generation $\mathcal{D}_{Rw}$ with zero-shot LLM rewriting. Specifically, we replace $\mathcal{D}_{Rw}$ by  $\bigcup_{s \in \mathbb{S}} \{ (\mathcal{P}_i, \mathcal{M}_\theta^0(\mathcal{P}_{ij} \vert \mathcal{P}_i) )_{1\leq j \leq 4}  \}_{\mathcal{P}_i \in \mathbb{P}^s} $, with $4$ rewrites per EHR (w/o Algorithm 1).
    \item \textit{w/o Rewriter}: we use an off-the-shelf LLM, $\mathcal{M}_\theta^0$, instead of the $\mathcal{M}_\theta$ EHR rewriter  (w/o step 1, Fig. \ref{fig:training}).
    \item \textit{w/o KL}: we train the LLM rewriter without the KL divergence loss, removing the alignment between the rewriter and the clinical prediction objectives (w/o step 3, Fig. \ref{fig:training}).

\end{enumerate}

We report our results in Table \ref{tab:ablation}. Consistent with our main results, Llama3-8B demonstrates marginally superior performance compared to Qwen2.5-7B across all the ablation scenarios. We can observe that ablating the EHR alignment component (\textit{w/o KL}) causes the most important performance degradation across both backbone LLMs, with the most pronounced effects on the imbalanced MOR prediction task. Specifically, we can see that substantial PRC drops of $31.5\%$ and $41.8\%$ for Qwen and Llama, respectively. Similarly, for the RA task, we observe a degradation between $4.7\%$ and $6.3\%$. For the LOS task, we observe a more modest degradation between $2.4\%$ and $3.6\%$, suggesting that KL training provides critical value, particularly for tasks with severe class imbalance.

Ablating the fine-tuning of the EHR rewriter (\textit{w/o Rewriter}) follows similar degradation patterns, with the MOR task suffering from a decrease of $31.5\%$ and $20.5\%$, for Qwen and Llama in PRC scores, respectively. This proves that fine-tuning the EHR rewriter is an equally crucial step in our pipeline.

Finally, we observe that removing our  synthetic training dataset (\textit{w/o $\mathcal{D}_{Rw}$}) has the most pronounced impact on the LOS task, indicating that pseudo-label augmentation provides greater value for tasks with larger training sets where diverse synthetic examples can capture broader clinical patterns.

\subsection{Model transferability}
We evaluate the transferability of the \textbf{ReToP} framework on unseen data, namely, eICU\footnote{We defined our clinical task at the ICU visit instead of the hospital visit following \cite{pyhealth2023yang}} and unseen task, namely,  MALF with the EFEMERIS dataset. We use the rewriter of each corresponding task\footnote{For MALF, we select the best EHR rewriter among the 3 tasks.}.
We adopt a low-cost adaptation approach,  implementing efficiency measures at both model and data levels.
Rather than fine-tuning the complete \textbf{ReToP} framework ($>8B$ params.),  we train only the clinical predictor with $149M$ params. ($1.8\%$ of total).
For data efficiency, we limit rewritten data to the test set and $4000$ training samples (equivalent to $3.6\%$ of the largest dataset) for predictor inoculation.
We use our best baseline classifier (ModernBERT). Due to computational constraints, we evaluate on $20\%$ of the test set for LOS and MALF tasks. As shown in Table \ref{tab:generalize_results}, ReToP$_{Qwen}$ and ReToP$_{Llama}$ improve PRC scores up to $+1.97\%$, $+0.46\%$, and  $+9.94\%$ for MOR, RA, and MALF tasks, correspondingly, while maintaining competitive LOS performance. These results demonstrate generalization with minimal adaptation cost, particularly for highly imbalanced tasks (MOR, RA, and MALF) with up to $3.05\%$ positive examples.

\begin{table}[tb]
    \centering
    \small
     \setlength{\tabcolsep}{2pt}
    \caption{Transferability performance with ROC/PRC metrics.} 
    \begin{tabular}{ccccc}
    \toprule
    \textbf{Model} &  \multicolumn{3}{c}{\textbf{MIMIC-IV $\xrightarrow{}$eICU}} &  \multicolumn{1}{c}{\textbf{$\xrightarrow{}$EFEMERIS}}\\
    \midrule
         &  \textbf{MOR} & \textbf{RA} &  \textbf{LOS} & \textbf{MALF}\\
    \midrule
    \small ModernBERT & $81.30$/$21.86$ & $86.71$/$23.80$&  $\mathbf{93.19}$/$56.67$ &$49.79$/$6.44$\\ 
    ReToP$_{Llama}$ & $\mathbf{81.42}$/$\mathbf{22.29}$ & $86.72$/$23.78$ & $93.16$/$56.78$& $\mathbf{51.53}$/$6.93$\\
    ReToP$_{Qwen}$ & $81.40$/$21.93$ & $\mathbf{86.79}$/$\mathbf{23.91}$ & $93.17$/$\mathbf{56.80}$ & $50.41$/$\mathbf{7.08}$  \\
    \bottomrule
    \end{tabular}    
    \label{tab:generalize_results}    
\end{table}

\subsection{Model analysis}

\begin{figure}[tb]
    \centering
    \includegraphics[width=0.85\columnwidth]{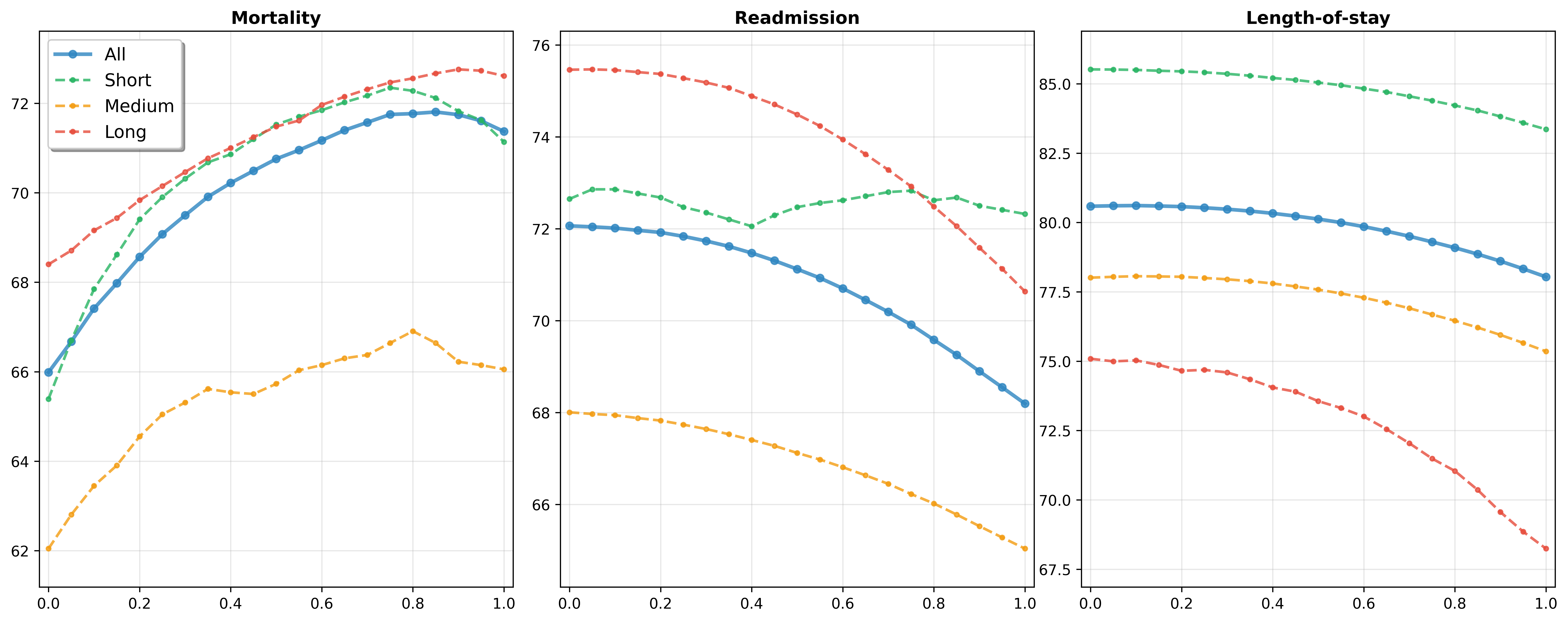}
    \caption{ ROC scores across  $\alpha$ values for inference level (§ Eq. 2) using ReToP$_{Llama}$ on MIMIC-IV, shown for MOR (left), RA (center), and LOS (right) tasks. Curves are stratified by EHR length: overall (blue), short EHR $<2048$ (green), medium EHR in $[2048,4096]$ (yellow), and long $>4096$ (red).}
    \Description{Three figures, one per task. Each figure has 4 ROC curves, each one representing a different EHR length.}
    \label{fig:alpha_values}
\end{figure}

\paragraph{Effect of interpolated prediction}
We evaluate the impact of the interpolation parameter $\alpha$ (§ Eq. 2) for leveraging original EHRs and rewrites at the inference stage. Figure \ref{fig:alpha_values} shows ROC performance across different $\alpha$ values (blue curves), where $\alpha=0$ represents the inference with no rewrites, and $\alpha=1$, using the EHR rewriter. Aligned with our previous findings, optimal values are task dependent. MOR benefits more from rewrite-only input, showing that our EHR rewriter mitigates sparsity and noise characteristics of mortality-related EHR data. 
Conversely, RA and LOS benefit from no rewrites ($\alpha=0$), suggesting that preserving full clinical detail is more valuable than noise reduction. 
Figure \ref{fig:alpha_values} further stratifies performance by EHR length. For the MOR task, rewrites ($\alpha=1$) consistently outperforms across all the input lengths. However, for RA and LOS, short EHRs show little differences between original and rewritten versions, while medium and longer EHRs exhibit degraded performance when using rewrites, suggesting that  rewrite quality decreases with increasing EHR length for these tasks.

\begin{figure}[tb]
    \centering
    \includegraphics[width=\columnwidth]{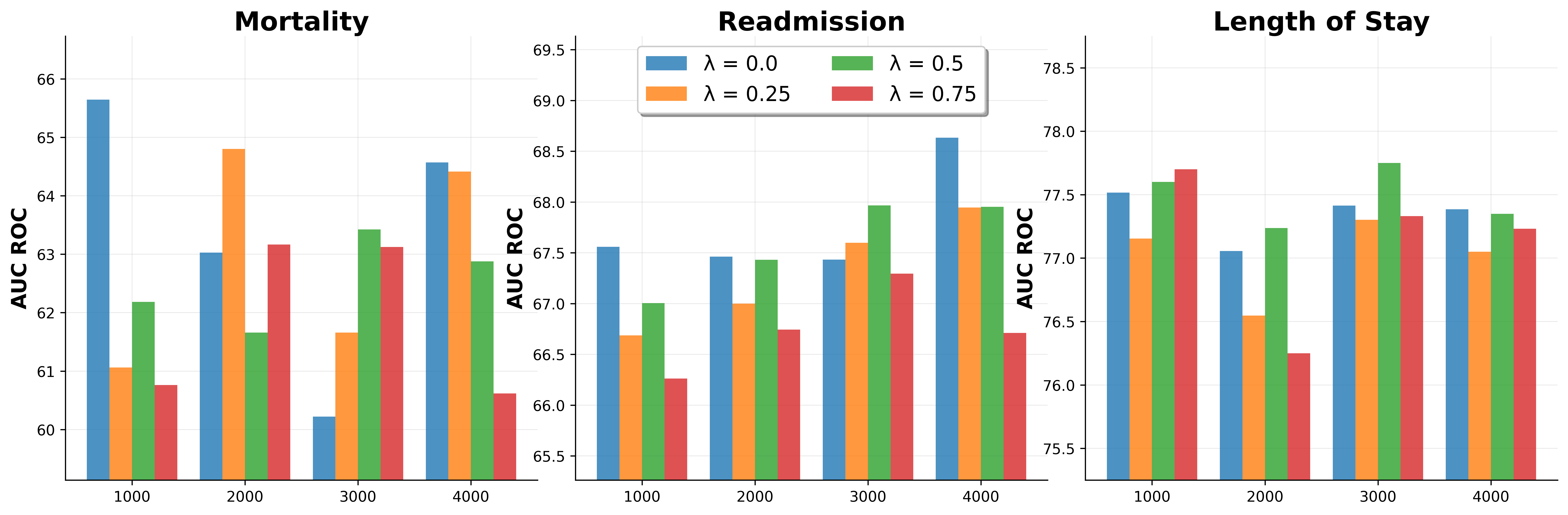}
     \caption{Effect of $\lambda$ on KL training (up to 4k  steps, eval. every 1000 steps) for MOR, RA, LOS using ReToP$_{Llama}$.}
         \Description{Three figures, one per task. Each figure 4 points with 4 bars. each bar represents a different lambda value.}
    \label{fig:training_KL}
\end{figure}

\paragraph{Effect of KL training}
We analyze KL divergence training across training steps and balancing parameter $\lambda$ (§ Eq. 10) using ReToP$_{Llama}$ without inoculation. 
Figure \ref{fig:training_KL} shows that that models with $\lambda \geq 0.5$ do not further improve after $3000-4000$ steps,  suggesting early stopping. For $\lambda \leq 0.25$, training is noisy with early overfitting except for RA. MOR is more sensitive to $\lambda$, benefiting from stronger regularization early on, while RA and LOS remain robust across $\lambda$ values. Our results suggest regularization should consider class imbalance.

\section{Case Study}
Our aim here is to check and then get insights into the faithfulness and clinical value under the expert perspective of the best \textbf{ReToP} setting. To achieve this goal, we  qualitatively analyze a sample of $15$ patient EHRs where \textbf{ReToP} accurately predicts the MOR task. For each case, we collect rewrites from \textit{self-gen}, \textit{ReToP w/o KL}, and full \textit{ReToP} using Llama backbone model. Annotators blindly evaluate each rewrite on two criteria using three labels ('Yes', 'Partially', 'No'): (1) \textit{Faithfulness}: whether the rewrite is entailed by the original EHR, judged by three reviewers including one expert, and (2) \textit{Actionability}: whether the rewrite includes clinical features that likely support the clinical decision-making, annotated by one expert. 

Table \ref{tab:qualitative_analysis} shows the results in terms of ratios for each label and each model on \textit{Faithfulness} and \textit{Actionability} criteria.
Overall, \textit{ReToP w/o KL} shows higher faithfulness than \textbf{ReToP} and \textit{self-gen} (based on the 'Y' annotation) and lower faithfulness after KL, leading to a decrease of the 'Y' and an increase of the 'P' and 'N' annotations. This trend is consistent with the expert annotation on the actionability, showing an opposite pattern on \textit{Actionability} criteria. We can see that \textit{self-gen} is more likely to include expected predictive features by the experts than the \textit{w/o KL} model and, more importantly, the full \textbf{ReToP} scenario, with nearly half of the annotations ($46$) revealing the increasing presence of unexpected features for the expert. All these results bring two important insights of the \textbf{ReToP} framework: (1) the EHR rewriter (w/o KL) intrinsically outputs EHR rewrites that are faithful to the original ones, as targeted in addressing challenge C1; (2) the KL alignment between the EHR rewriter and clinical predictors revises the rewriter by emphasizing predictive clinical features, in addressing C2, but which are seemingly serendipitous for the expert regarding the clinical task at hand. 

\begin{figure}[tb]
    \footnotesize
    \centering
    \includegraphics[trim=0 1.5cm 12cm 0, clip, width=0.8\columnwidth]{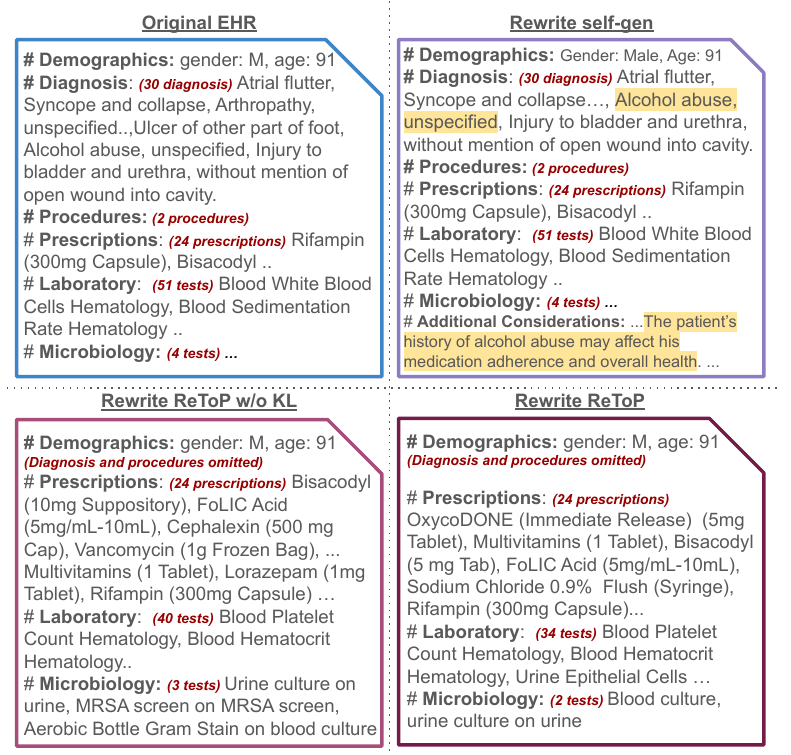}
    \caption{Example patient EHR for the MOR task. Red text shows main content-based differences across rewrites, while \textbf{highlighted text} shows generated unfaithful information.}
    \Description{Four different text versions of a patient profile. Each version correspond to a different text-based generation for the self-gen model, and ReToP with and without KL.}
    \label{fig:examples}
\end{figure}

We dig into these results with a qualitative analysis on Figure \ref{fig:examples}.
Examining \textit{faithfulness} across all the rewrites, \textit{self-gen} preserves all the original features but introduces unfaithful information (highlighted in yellow). For instance, while the original EHR indicates ``Alcohol abuse, unspecified, the \textit{self-gen} rewrite adds complementary information stating that ``... alcohol abuse may impact his medication, not present in the source data.
In contrast, both \textbf{ReToP}-based rewrites lean to filter features rather than adding new spans, which preserves their \textit{faithfulness}.  Interestingly, we can see that \textit{ReToP w/o KL} reduces diagnoses from $30$ to $24$ and laboratory tests from $51$ to $40$, while \textbf{ReToP} further reduces tests to $34$, retaining only task-relevant features that optimize prediction. This would explain  the decrease in  \textit{actionability} according to the expert perspective and opens avenues of research for model explanability about the underlying reasons to filter.

\begin{table}[tbp]
    \centering
    \small
    \caption{Qualitative analysis of \textbf{ReToP} (w/o and w/ KL models) vs. self-gen  on MOR based on \textit{faithfulness} and \textit{actionability}.}
    \begin{tabular}{ccccccc}
    \toprule
    \textbf{Model} & \multicolumn{3}{c}{\textbf{\% Faithfulness}} & \multicolumn{3}{c}{\textbf{\% Actionability}} \\
    & Y &P &N & Y &P &N  \\
    \midrule
    self-gen & 60 & 40 & 0 & 80&20&0 \\
    ReToP (w/o KL) & 73 &27&0&0&100&0 \\
    ReToP & 71 & 22 & 7 & 7 & 47 & 46 \\
    \bottomrule
    \end{tabular}
    \label{tab:qualitative_analysis}
\end{table}

\section{Conclusion}\label{conclusion}
We introduced \textbf{ReToP}, a new  framework that leverages LLMs to enhance clinical prediction performance. \textbf{ReToP} trains an LLM-based EHR rewriter using synthetic EHR rewrites built upon health-related feature selection methods. Then, \textbf{ReToP} refines the EHR rewriter through an end-to-end training guided by the clinical predictor supervision using a KL loss. \textbf{ReToP} significantly outperforms a set of state-of-the-art baselines across representative clinical prediction tasks.  Our proposed framework exhibits reasonable transfer ability to out-of-domain datasets and tasks. By designing EHR rewriters that can be efficiently aligned with downstream clinical tasks,   \textbf{ReToP} opens up potential directions for effective healthcare AI systems. Future work could explore extending this framework to a wider range of clinical tasks, including multi-label classification tasks, and investigating the right compromise between model performance and model explainability for domain experts.

\begin{acks}  
This work has been supported by the In-Utero project funded by HDH (France) and FRQS (Canada). This work was also granted access to the HPC resources of IDRIS under the allocation 2025-AD011015371R1 made by GENCI.

\section*{Ethical Considerations}
In this work, we used available de-identified datasets from the medical domain, including MIMIC-IV, eICU, and EFEMERIS datasets, with proper attribution to sources. These datasets contain patient data that has been anonymized and de-identified by established privacy protection standards, ensuring no individual patient information can be traced or identified.

While our \textbf{ReToP} framework demonstrates improved clinical prediction performance, we emphasize that these predictions should be exclusively used as decision-supporting tools for experts. 
\end{acks}


\bibliographystyle{ACM-Reference-Format}
\balance 
\bibliography{general}

\end{document}